\documentclass[journal]{IEEEtran}
\usepackage{amssymb,latexsym,graphicx,times,amsmath,subfigure,threeparttable,booktabs,bm,color,multirow,cite}

\makeatletter

\newcommand{\Rmnum}[1]{\expandafter\@slowromancap\romannumeral #1@}
\makeatother

\begin{document}
\title{Adversarial Attacks and Defenses\\ in Physiological Computing: A Systematic Review}

\author{Dongrui~Wu, Jiaxin Xu, Weili~Fang, Yi~Zhang, Liuqing~Yang,  Xiaodong~Xu, Hanbin~Luo and Xiang~Yu

\thanks{D.~Wu and J.~Xu are with the Ministry of Education Key Laboratory of Image Processing and Intelligent Control, School of Artificial Intelligence and Automation, Huazhong University of Science and Technology, Wuhan 430074 China. D. Wu is also with Zhejiang Lab, Hangzhou 311121 China. Email: drwu@hust.edu.cn, jiaxinxu@hust.edu.cn. }

\thanks{W.~Fang and H.~Luo are with the School of Civil and Hydraulic Engineering, Huazhong University of Science and Technology, Wuhan 430074 China. Email: weili\_f@hust.edu.cn, luohbcem@hust.edu.cn.}

\thanks{Y.~Zhang and X.~Xu are with the College of Public Administration, Huazhong University of Science and Technology, Wuhan 430074 China. Email: yizhanghn@sina.com, xiaodong-xu@hust.edu.cn.}

\thanks{L.~Yang is with the University of Michigan, Ann Arbor, MI 48109 USA. Email: yanglq@umich.edu.}

\thanks{X.~Yu (Member of the Academy of Europe) is with the School of Management and Sino-European Institute for Intellectual Property, Huazhong University of Science and Technology, Wuhan 430074 China. Email: yuxiang@hust.edu.cn.}

\thanks{X.~Xu, H.~Luo and X.~Yu are the corresponding authors.}
}

\markboth{National Science Open}
{Wu \MakeLowercase{\textit{et al.}}: Adversarial Attacks and Defenses in Physiological Computing: A Systematic Review}
\maketitle

\begin{abstract}
Physiological computing uses human physiological data as system inputs in real time. It includes, or significantly overlaps with, brain-computer interfaces, affective computing, adaptive automation, health informatics, and physiological signal based biometrics. Physiological computing increases the communication bandwidth from the user to the computer, but is also subject to various types of adversarial attacks, in which the attacker deliberately manipulates the training and/or test examples to hijack the machine learning algorithm output, leading to possible user confusion, frustration, injury, or even death. However, the vulnerability of physiological computing systems has not been paid enough attention to, and there does not exist a comprehensive review on adversarial attacks to them. This paper fills this gap, by providing a systematic review on the main research areas of physiological computing, different types of adversarial attacks and their applications to physiological computing, and the corresponding defense strategies. We hope this review will attract more research interests on the vulnerability of physiological computing systems, and more importantly, defense strategies to make them more secure.
\end{abstract}


\IEEEpeerreviewmaketitle

\section{Introduction}

Keyboards and mouses, and recently also touchscreens, are the most popular means that a user sends commands to a computer. However, they convey little information about the psychological state of the user, e.g., cognitions, motivations and emotions, which are also very important in the development of `smart' technology \cite{Fairclough2009}. For example, on emotions, Marvin Minsky, a pioneer in artificial intelligence, pointed out early in the 1980s that \cite{Minsky1988} ``\emph{the question is not whether intelligent machines can have any emotions, but whether machines can be intelligent without emotions.}"

Physiological computing \cite{Jacucci2015} is ``\emph{the use of human physiological data as system inputs in real time.}" It opens up bandwidth within human-computer interaction by enabling an additional channel of communication from the user to the computer \cite{Fairclough2009}, which is necessary in adaptive and collaborative human-computer symbiosis.

\begin{figure}[htpb] \centering
\includegraphics[width=.5\linewidth,clip]{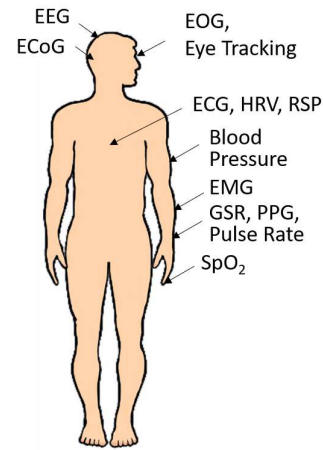}
\caption{Common signals in physiological computing, and their typical measurement locations.} \label{fig:location}
\end{figure}

Common physiological data in physiological computing include the electroencephalogram (EEG), electrocorticogram (ECoG), electrocardiogram (ECG), electrooculogram (EOG), electromyogram (EMG), eye movement, blood pressure (BP), electrodermal activity (EDA), respiration (RSP), skin temperature, etc., which are recordings or measures produced by the physiological process of human beings. Their typical measurement locations are shown in Fig.~\ref{fig:location}.

These signals have been widely studied in the literature in various applications, including clinic diagnostics, and wearable devices for health monitoring and human-machine interactions, as indicated by the number of publications in Table~\ref{tab:pubs}. The top four most frequently studied physiological signals are blood pressure, EEG, ECG and respiration. Blood pressure and respiration are both vital signs. Multi-lead ECG has been widely used in hospitals for screening and diagnosis of cardiovascular diseases, and single-lead ECG has been incorporated into millions of smart watches and wristbands for fitness tracking and atrial fibrillation early warning \cite{Han2020}. EEG is popular may be because it is the most frequently used input signal in brain-computer interfaces (BCIs) \cite{Lance2012} for neural rehabilitation \cite{Daly2008}, consciousness evaluation \cite{Huang2019}, emotion regularization \cite{Shanechi2019}, text input \cite{Chen2015a},  external device control \cite{Wolpaw2002}, etc., and gold standard in certain clinic practices, e.g., seizure diagnostics \cite{drwuSeizure2022}.

\begin{table}[htpb] \centering \setlength{\tabcolsep}{1.5mm}
\caption{Common signals in physiological computing, and the corresponding number of publications in Google Scholar with the keywords in title (as of 12/24/2021).} \label{tab:pubs}
\begin{tabular}{p{2cm} p{4cm} c}\toprule
Signal & Keywords           & No. Publications   \\ \midrule
Electro-encephalogram  & EEG OR Electroencephalogram OR Electroencephalography   & 155,000  \\
Electrocardiogram     & ECG OR EKG OR Electrocardiogram       & 104,000\\
Electromyogram        & EMG OR Electromyogram          & 44,900\\
Electrocorticogram & ECoG OR Electrocorticogram OR Electrocorticography &4,340\\
Electrooculogram      & EOG OR Electrooculogram & 2,690\\
Respiration     &  Respiration & 92,300\\
Blood Pressure        & Blood Pressure               &217,000 \\
Heart Rate Variability& HRV OR Heart Rate Variability  & 40,400\\
Electrodermal Activity & EDA OR GSR OR EDR OR Electrodermal OR Galvanic Skin Response    & 17,800 \\
Eye Movement          & Eye Movement OR Eye Tracking     & 16,900\\
Oxygen Saturation     &  SpO2 OR Oxygen Saturation OR Blood Oxygen & 26,800\\
Skin Temperature      & Skin Temperature & 7,320\\
Photo-plethysmogram    & PPG OR Photoplethysmogram  & 5,390 \\
Pulse Rate            &Pulse Rate & 7,160\\  \bottomrule
\end{tabular}
\end{table}

\begin{figure}[htpb] \centering
\includegraphics[width=.8\linewidth,clip]{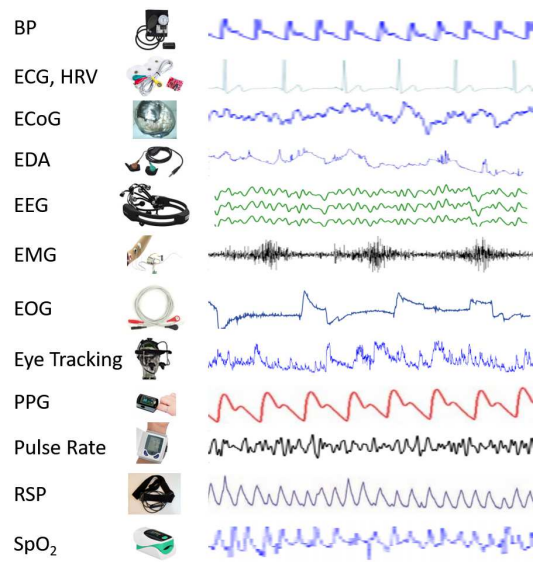}
\caption{Examples of common signals in physiological computing, and their typical measurement equipment. Blood pressure is usually measured by auscultation. ECG/HRV, ECoG, EDA, EEG, EMG and EOG are bioelectrical signals generated by nerves and muscles. Eye tracking, PPG and SpO$_2$ are bio-optical signals.} \label{fig:signals}
\end{figure}

Physiological signals are usually single-channel or multi-channel time series, as shown in Fig.~\ref{fig:signals}. In many clinical applications, the recording may last hours, days, or even longer. For example, long-term video-EEG monitoring for seizure diagnostics may need 24 hours, and ECG monitoring in intensive care units (ICUs) may last days or weeks. Wearable ECG monitoring devices, e.g., iRythm Zio Patch, AliveCor KardiaMobile, Apple Watch, and Huawei Band, are being used by millions of users. A huge amount of physiological signals are collected during these processes. Manually labelling them is very labor-intensive, and even impossible for wearable devices, given the huge number of users.

Machine learning \cite{drwuTLBCI2022} has been used to alleviate this problem, by automatically classifying the measured physiological signals. Particularly, deep learning has demonstrated outstanding performances \cite{Rim2020}, e.g., EEGNet \cite{EEGNet}, DeepCNN \cite{Schirrmeister2017}, ShallowCNN \cite{Schirrmeister2017} and TIDNet \cite{Kostas2020} for EEG classification, SeizureNet for EEG-based seizure recognition \cite{Asif2020}, CNN for ECG rhythm classification \cite{Goodfellow2018}, ECGNet for ECG-based mental stress monitoring \cite{Hwang2018a}, and so on.

However, recent research has shown that both traditional machine learning and deep learning models are vulnerable to various types of attacks \cite{Szegedy2014,Goodfellow2015,Qiu2019,Miller2020}. For example, Sharif \emph{et al.} \cite{Sharif2016} successfully fooled a face recognition system using a deliberately designed eyeglass for the target face. Brown \emph{et al.} \cite{Brown2017} generated adversarial patches, which could be placed anywhere within an image and caused the classifier to output the target class.
Chen \emph{et al.} \cite{Chen2017} created a backdoor in the target model by injecting poisoning samples, which contain an ordinary sunglass, into the training set, so that all test images with the sunglass would be classified into a target class. Athalye \emph{et al.} \cite{Athalye2018} synthesized a 3D adversarial turtle, which was classified as a rifle at every viewpoint. Eykholt \emph{et al.} \cite{Evtimov2018} stuck a carefully crafted graffiti to road signs, and caused the model to classify `Stop' as `Speed limit 40'. Finlayson \emph{et al.} \cite{Finlayson2018,Finlayson2019} successfully performed adversarial attacks to deep learning classifiers across three clinical domains (fundoscopy, chest X-ray, and dermoscopy). Rahman \emph{et al.} \cite{Rahman2021} performed adversarial attacks to six COVID-19 related applications, including recognizing whether a subject is wearing a mask, maintaining deep learning based QR codes as immunization certificates, recognizing COVID-19 from CT scan or X-ray images, etc. Ma \emph{et al.} \cite{Ma2020} showed that medical deep learning models can be more vulnerable to adversarial attacks than models for natural images, but surprisingly and fortunately, medical adversarial attacks may also be easily detected. Kaissis \emph{et al.} \cite{Kaissis2020} pointed out that various other attacks, in addition to adversarial attacks, also exist in medical imaging, and called for secure, privacy-preserving and federated machine learning to cope with them.

Machine learning models in physiological computing are not exempt from adversarial attacks \cite{drwuNSR2021,Karimian2019,Karimian2020}. However, to the best of our knowledge, there does not exist a systematic review on adversarial attacks in physiological computing. This paper fills this gap, by comprehensively reviewing different types of adversarial attacks, their applications in physiological computing, and possible defense strategies. It will be very important to the security of physiological computing systems in real-world applications.

We need to emphasize that this paper focuses on the emerging adversarial attacks and defenses. For other types of attacks and defenses, e.g., cybersecurity, the readers can refer to, e.g., \cite{Bernal2020}.

The remainder of this paper is organized as follows: Section~\ref{sect:PC} introduces five relevant research areas in physiological computing. Section~\ref{sect:types} introduces different categorizations of adversarial attacks. Section~\ref{sect:attacks} describes various adversarial attacks to physiological computing systems. Section~\ref{sect:defense} introduces different approaches to defend against adversarial attacks, and their applications in physiological computing. Finally, Section~\ref{sect:conclusion} draws conclusions and points out some future research directions.

\section{Physiological Computing} \label{sect:PC}

As physiological computing uses human physiological data as system inputs in real time, it includes, or significantly overlaps with, BCIs, affective computing, adaptive automation, health informatics, and physiological signal based biometrics. Although these five areas have different application scenarios and goals, they all need to build machine learning models for physiological signal classification or regression, and hence all are subject to adversarial attacks.

\subsection{BCIs}

A BCI system establishes a direct communication pathway between the brain and an external device, e.g., a computer or a robot \cite{Lance2012}. Scalp and intracranial EEGs have been widely used in BCIs \cite{drwuTLBCI2022}.

\begin{figure}[htbp]         \centering
\includegraphics[width=\linewidth,clip]{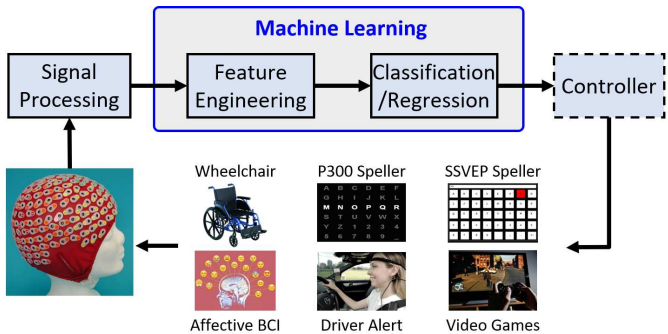}
    \caption{Flowchart of a closed-loop EEG-based BCI system. } \label{fig:BCI}
\end{figure}

The flowchart of a closed-loop EEG-based BCI system is shown in Fig.~\ref{fig:BCI}. After EEG signal acquisition, signal processing, usually including both temporal filtering and spatial filtering, is used to enhance the signal-to-noise ratio. Machine learning is next performed to understand what the EEG signal means, based on which a control command may be sent to an external device.

There are three typical paradigms in EEG-based BCIs \cite{drwuTLBCI2022}:
\begin{enumerate}
\item \emph{Motor imagery (MI)} \cite{Pfurtscheller2001}, which modifies neuronal activities in primary sensorimotor areas when the user imagines the movement of various body parts, e.g., the left (right) hemisphere for right-hand (left-hand) MIs and center for feet MIs. These MIs can be decoded to control external devices, e.g., a wheelchair, or in neural rehabilitation \cite{Daly2008} to restore the functionality of hands after stroke.
\item \emph{Event-related potentials (ERPs)} \cite{Handy2005,Lees2018}, which are stereotyped EEG responses to rare or expected visual, audio, or tactile stimuli. The most frequently used ERP component is P300 \cite{Sutton1965}, which is an increase of the EEG magnitude observed about 300 ms after a rare stimulus.
\item \emph{Steady-state visual evoked potential (SSVEP)} \cite{Friman2007}, which is brain's electrical response to repetitive visual stimulation, usually between 3.5 and 75 Hz \cite{Beverina2003}. SSVEP can achieve very high information transfer rate in BCI spellers \cite{Chen2015a}.
\end{enumerate}

EEG-based BCI spellers may be the only non-muscular communication devices for Amyotrophic Lateral Sclerosis (ALS) patients to express their opinions \cite{Sellers2006}. In seizure treatment, responsive neurostimulation (RNS) \cite{Geller2018,Gummadavelli2018} recognizes ECoG or intracranial EEG patterns prior to ictal onset, and delivers a high-frequency stimulation impulse to stop the seizure, improving the patient's quality-of-life.

\subsection{Affective Computing}

Affective computing \cite{Picard1997} is ``\emph{computing that relates to, arises from, or deliberately influences emotion or other affective phenomena}."

\begin{figure}[htpb] \centering
\includegraphics[width=.6\linewidth,clip]{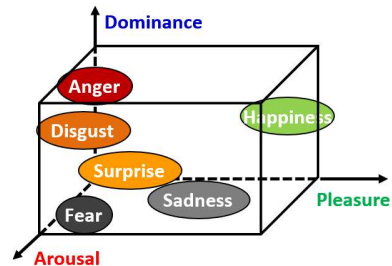}
\caption{Ekman's six basic emotions in the 3D space of Arousal, Pleasure and Dominance.} \label{fig:3D}
\end{figure}

Emotion is the focus of affective computing. It can be represented by discrete categories, e.g., Ekman's six basic emotions \cite{Ekman1971} (Anger, Disgust, Fear, Happiness, Sadness, and Surprise), and by continuous values in the 2D space of Arousal and Pleasure (or Valence) \cite{Russell1980}, or the 3D space of Arousal, Pleasure (or Valence) and Dominance \cite{Mehrabian1980}, as shown in Fig.~\ref{fig:3D}.

Various inputs can be used in affective computing, e.g., videos, text, speech, etc. Physiological signals have also been extensively used \cite{drwuAC2021}. In bio-feedback based relaxation training \cite{Chittaro2014}, EDA can be used to detect the user's affective state, based on which a relaxation training application can provide the user with explicit feedbacks to learn how to change his/her physiological activities to improve health and performance. In software adaptation \cite{Aranha2021}, the graphical interface, difficulty level, sound effects and/or contents are automatically adapted based on the user's real-time emotion estimated from various physiological signals, to keep the user more engaged.

\subsection{Adaptive Automation}

Adaptive automation controls the number and/or types of tasks allocated to the operator to keep the workload within an appropriate level (avoiding both underload and overload), and hence to enhance the overall performance and safety of the human-machine system \cite{Boeke2015,Arico2016}.

\begin{figure}[htpb] \centering
\includegraphics[width=\linewidth,clip]{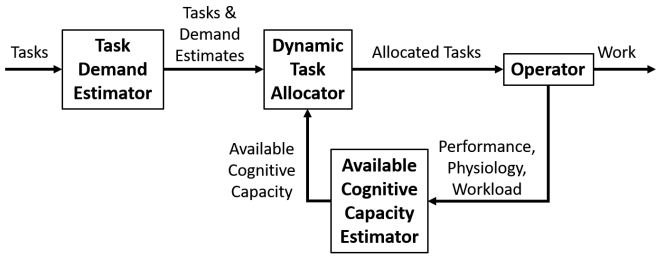}
\caption{Adaptive automation expressed as a control system \cite{Boeke2015}.} \label{fig:AA}
\end{figure}

Boeke \emph{et al.} \cite{Boeke2015} expressed adaptive automation as a control system, shown in Fig.~\ref{fig:AA}. The task demand estimator maps each task to a cognitive demand. The dynamic task allocator allocates tasks to the operator based upon his/her available cognitive capacity and the incoming tasks' cognitive demands. The available cognitive capacity estimator estimates the operator's available cognitive capacity from his/her performance, physiology, or subjective measures.

In air traffic management \cite{Arico2016}, an operator's EEG signal can be used to estimate the mental workload, and trigger specific adaptive automation solutions. This can significantly reduce the operator's workload during high-demanding conditions, and increase the task execution performance. A study \cite{Greef2009} also showed that pupil diameter and fixation time, measured from an eye-tracking device, can be indicators of mental workload, and hence be used to trigger adaptive automation.

Park and Zahabi \cite{Park2022} performed a review on cognitive workload assessment of prosthetic devices, which can be achieved using physiological, subjective, or task performance measures. The first includes EEG, EMG, ECG, EDA, respiration, eye-tracking, etc. They found that hybrid inputs, e.g., EMG plus inertial measurement unit (IMU), or EMG plus force myography (FMG), were less cognitively demanding than EMG or EEG alone. More specifically, the combination of EMG and IMU can improve the effectiveness, satisfaction and efficiency of prosthetic devices, and the combination of EMG and FMG achieved higher overall stability (lower variance) than EMG alone.

\subsection{Health Informatics}

Health informatics studies information and communication processes and systems in healthcare \cite{Coiera2015}.

A single-lead short ECG recording (9-60 seconds), collected from the AliveCor personal ECG monitor, can be used by a convolutional neural network (CNN) to classify normal sinus rhythm, atrial fibrillation, an alternative rhythm, or noise, with an average test accuracy of 88\% on the first three classes \cite{Han2020}. A recent study \cite{Mishra2020} also showed that heart rate data from consumer smart watches, e.g., Apple, Fitbits and Garmin devices, can be used for pre-symptomatic detection of COVID-19, sometimes nine or more days earlier.

\begin{figure}[htpb] \centering
\includegraphics[width=\linewidth,clip]{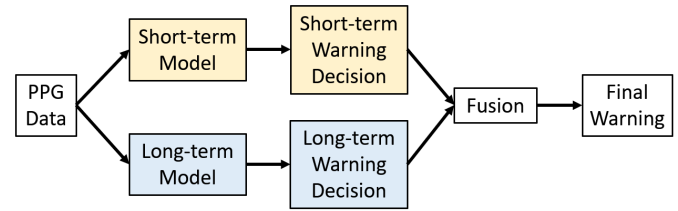}
\caption{A model fusion approach for PPG-based atrial fibrillation onset prediction \cite{Guo2021}.} \label{fig:AF}
\end{figure}

Many smart watches or wristbands record the PPG signal, a measure of arterial blood volume fluctuating with
each heartbeat \cite{Charlton2022}. It is frequently used to monitor the heart rate, but also contains information on the cardiac, vascular, respiratory, and autonomic nervous systems. In clinics, wearable PPGs can be used for atrial fibrillation detection, obstructive sleep apnea identification, monitoring the spread of infectious diseases, sleep monitoring, mental stress assessment, vascular age assessment, clinical deterioration identification, cardiovascular risk prediction, response to exercise assessment, sepsis identification, heart failure identification, and preeclampsia identification \cite{Charlton2022}. For example, Guo \emph{et al.} \cite{Guo2021} proposed a model fusion approach (Fig.~\ref{fig:AF}) for PPG-based atrial fibrillation onset prediction, using Huawei smart watches and wristbands. It achieved 94.04\% sensitivity, 96.35\% specificity, and 94.04\% recall.

\subsection{Physiological Signal Based Biometrics}

Physiological signal based biometrics \cite{Singh2012} use physiological signals for biometric applications, e.g., digitally identify a person to grant access to systems, devices or data. They represent a paradigm shift from conventional ``something we know" (e.g., a personal identification number) or ``something we have" (e.g., an access card) policies to ``something we are" \cite{Thomas2017}.

\begin{figure}[htpb] \centering
\includegraphics[width=\linewidth,clip]{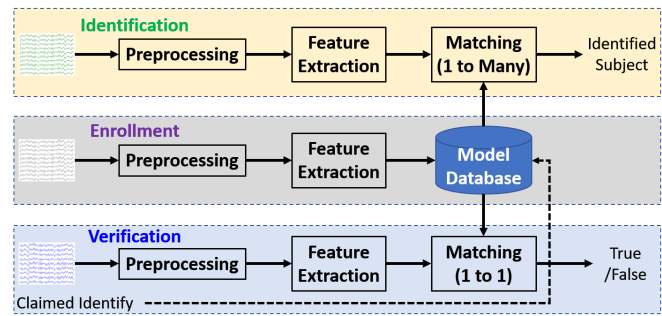}
\caption{Enrollment, identification and authentication in biometrics \cite{Thomas2017}.} \label{fig:Biometrics}
\end{figure}

A biometric system typically includes three modes \cite{Thomas2017}: enrollment, identification, and authentication, as shown in Fig.~\ref{fig:Biometrics}. The enrollment mode converts each subject's physiological signals into a feature template and stores it in a database. The identification mode finds the best possible match for an incoming subject's template. The authentication mode verifies if a subject is indeed the person he/she claims to be.

EEG \cite{Thomas2017}, ECG \cite{Agrafioti2011}, PPG \cite{Yadav2018} and multimodal physiological signals \cite{Bianco2019} have been used in user identification and authentication, with the advantages of universality, permanence, liveness detection, continuous authentication, etc. Thomas and Vinod \cite{Thomas2017} performed a review on EEG-based biometric systems, and found that EEGs from the resting state with eyes closed or open, motor imagination, visual evoked potentials, and mental tasks (e.g., math operation, letter composition) can all be used in biometrics. Agrafioti \emph{et al.} \cite{Agrafioti2011} gave a comprehensive introduction of the theory, methods and applications of heart biometrics, pointing out their challenges, including time dependency, collection periods, privacy implications, and cardiac conditions. Bianco and Napoletano \cite{Bianco2019} performed multimodal biometric recognition using heart rate, breathing rate, palm EDA, and perinasal perspitation, achieving 90.54\% top-1 accuracy and 99.69\% top-5 accuracy.

\section{Adversarial Attacks} \label{sect:types}

\begin{figure}[htpb] \centering
\includegraphics[width=.8\linewidth,clip]{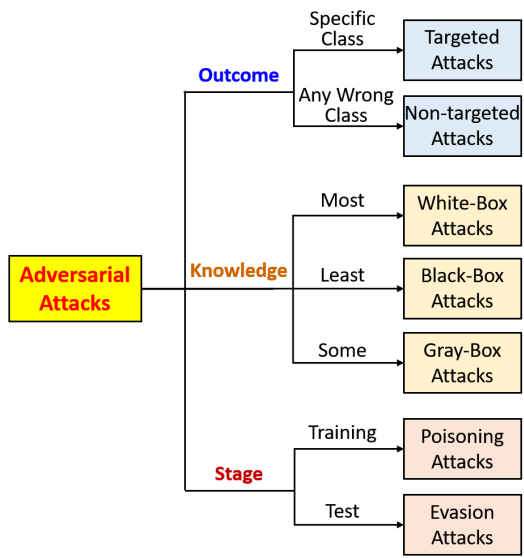}
\caption{Types of adversarial attacks.} \label{fig:types}
\end{figure}

Adversarial attacks generate various adversarial perturbations, which may be unnoticeable by human eyes or a computer program, to fool a machine learning model. There are different categorizations of adversarial attacks \cite{Qiu2019,Miller2020}, as shown in Fig.~\ref{fig:types}.

\subsection{Targeted and Non-targeted Attacks}

According to the outcome, there are two types of adversarial attacks \cite{Miller2020}: \emph{targeted attacks} and \emph{non-targeted (indiscriminate) attacks}.

Targeted attacks force a model to classify certain examples, or a certain region of the feature space, into a specific (usually wrong) class. Non-targeted attacks force a model to misclassify certain examples or feature space regions, but do not specify which class they should be misclassified into.

For example, in a 3-class classification problem, assume the class labels are $A$, $B$ and $C$. Then, a targeted attack may force the input to be classified into Class $A$, no matter what its true class is. A non-targeted attack forces an input from Class $A$ to be classified into Class $B$ or $C$, but does not specify it must be $B$ or $C$; as long as it is not $A$, then the non-targeted attack is successful.

\subsection{White-Box, Black-Box and Gray-Box Attacks}

According to how much the attacker knows about the target model, there can be three types of attacks \cite{drwuBCIAttack2019}:
\begin{enumerate}
    \item \emph{White-box attacks}, in which the attacker knows everything about the target model, including its architecture and parameters. This is the easiest attack scenario and could cause the maximum damage. It may correspond to the case that the attacker is an insider, or the model designer is evaluating the worst-case scenario when the model is under attack. Popular attack approaches include L-BFGS \cite{Szegedy2014}, DeepFool \cite{Moosavi-Dezfooli2016}, the C\&W method \cite{Carlini2017}, the fast gradient sign method (FGSM) \cite{Goodfellow2015}, the basic iterative method (BIM) \cite{Kurakin2017}, etc.

    \item \emph{Black-box attacks}, in which the attacker knows neither the architecture nor the parameters of the target model, but can supply inputs to the model and observe its outputs. This is the most realistic and also the most challenging attack scenario. One example is that the attacker purchases a commercial BCI system and tries to attack it. Black-box attacks are possible, due to the transferability of adversarial examples \cite{Szegedy2014}, i.e., an adversarial example generated from one machine learning model may be used to fool another machine learning model at a high success rate, if the two models solve the same task. So, in black-box attacks \cite{Papernot2017}, the attacker can query the target model many times to construct a training set, train a substitute machine learning model from it, and then generate adversarial examples from the substitute model to attack the original target model.

    \item \emph{Gray-box attacks}, which assume the attacker knows a limited amount of information about the target model, e.g., (part of) the training data that the target model is tuned on. They are frequently used in data poisoning attacks, as introduced in the next subsection.
\end{enumerate}
Table~\ref{tab:summary} compares the main characteristics of the three attack types.

\begin{table}[htbp] \centering \setlength{\tabcolsep}{2.5mm}
\caption{Comparison of white-box, gray-box and black-box attacks \cite{drwuBCIAttack2019}. `$-$' means that whether the information is available or not does not affect the attack strategy, since it is not used in the attack.}   \label{tab:summary}
\begin{tabular}{c|ccc}
\toprule
Target model information& White-Box & Gray-Box & Black-Box \\ \midrule
Know its architecture & $\checkmark$ & $\times$& $\times$ \\
Know its parameters &  $\checkmark$  & $\times$ & $\times$\\
Know its training data &  $-$ &  $\checkmark$&  $\times$ \\
Can observe its response &  $-$ & $-$& $\checkmark$ \\
\bottomrule
\end{tabular}
\end{table}

\subsection{Poisoning and Evasion Attacks}

\begin{figure}[htpb] \centering
\includegraphics[width=\linewidth,clip]{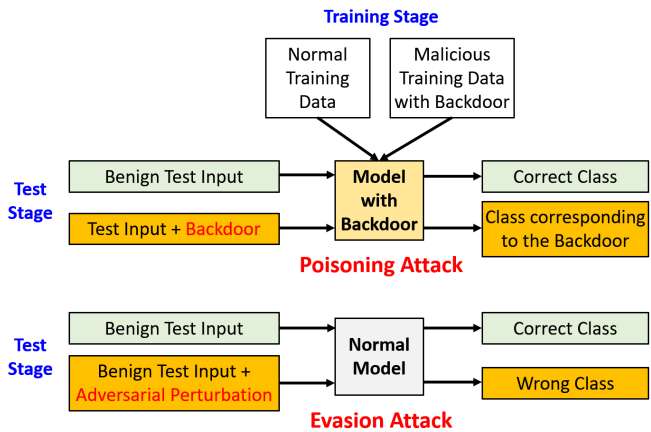}
\caption{Poisoning and evasion attacks.} \label{fig:attacks}
\end{figure}

According to the stage that the adversarial attack is performed, there are two types of attacks: \emph{poisoning attacks} and \emph{evasion attacks}, as shown in Fig.~\ref{fig:attacks}.

Poisoning attacks \cite{Xiao2015a} focus on the training stage, to create backdoors in the machine learning model by adding contaminated examples to the training set. At the test stage, an input with the backdoor can be classified into the class the attacker specifies. They are usually white-box or gray-box attacks, achieved by data injection, i.e., adding adversarial examples to the training set \cite{Mei2015}, or data modification, i.e., poisoning the training data by modifying their features or labels \cite{Biggio2011}.

Evasion attacks \cite{Goodfellow2015} happen at the test stage, by adding deliberately designed tiny perturbations to benign test examples to mislead the machine learning model. They are usually white-box or black-box attacks.

\section{Adversarial Attacks in Physiological Computing} \label{sect:attacks}

Most adversarial attack studies considered computer vision applications, where the inputs are 2D images. Physiological signals are continuous time series, which are quite different from images. There are relatively few adversarial attack studies on time series \cite{Fawaz2019,Karim2021,Harford2021,Cheng2022}, and even fewer on physiological signals. A summary of them is shown in Table~\ref{tab:comp1}.

\begin{table*}[htbp] \centering \setlength{\tabcolsep}{1.2mm}
\caption{Summary of existing adversarial attack approaches in physiological computing.}   \label{tab:comp1}
\begin{tabular}{c|c|c|cc|ccc|cc} \toprule
\multirow{2}{*}{Application}&\multirow{2}{*}{Problem}&\multirow{2}{*}{Reference}&\multicolumn{2}{|c}{Outcome} & \multicolumn{3}{|c|}{Knowledge} & \multicolumn{2}{c}{Stage}\\ \cline{4-10}
&& & Targeted & Non-targeted & White-box & Gray-box & Black-box & Poisoning & Evasion \\ \midrule
\multirow{6}{*}{BCI} &\multirow{6}{*}{Classification}& \cite{drwuBCIAttack2019} & &$\checkmark$&$\checkmark$&$\checkmark$&$\checkmark$& &$\checkmark$\\
&&\cite{drwuALBCI2019} & &$\checkmark$&&&$\checkmark$& &$\checkmark$\\
&&\cite{drwuNSR2021} & $\checkmark$ & & $\checkmark$ & &&&$\checkmark$\\
&&\cite{drwuUAP2021}& $\checkmark$ & $\checkmark$& $\checkmark$ &$\checkmark$ &&&$\checkmark$ \\
&&\cite{drwuEngineering2022} & $\checkmark$ & & &  & $\checkmark$ & $\checkmark$&\\
&&\cite{drwuSCIS2022} & $\checkmark$ & & & $\checkmark$ & & & $\checkmark$  \\ \cline{2-10}
&Regression& \cite{drwuTAR2019} &$\checkmark$ &&$\checkmark$&&$\checkmark$&&$\checkmark$  \\ \midrule

\multirow{5}{*}{Health Informatics} &\multirow{4}{*}{Classification}&  \cite{Han2020} &$\checkmark$ &$\checkmark$ &$\checkmark$&&&&$\checkmark$ \\
&& \cite{Aminifar2020} &$\checkmark$ & &$\checkmark$&&&&$\checkmark$ \\
&& \cite{Newaz2020} &$\checkmark$ &$\checkmark$ &$\checkmark$&&$\checkmark$&$\checkmark$&$\checkmark$\\
&&\cite{Wang2020} &$\checkmark$ &&$\checkmark$&&&$\checkmark$& \\
&&\cite{Karim2021} & &$\checkmark$&$\checkmark$&&$\checkmark$&&$\checkmark$ \\ \midrule

\multirow{4}{*}{Biometrics} &\multirow{4}{*}{Classification}&  \cite{Maiorana2013}
&$\checkmark$ &&&&$\checkmark$&&$\checkmark$ \\
&& \cite{Eberz2017} &$\checkmark$ &&&$\checkmark$&&&$\checkmark$ \\
&& \cite{Karimian2020} &$\checkmark$ &&&$\checkmark$&&&$\checkmark$ \\
&& \cite{Karimian2019} &$\checkmark$ &&&$\checkmark$&&&$\checkmark$\\ \bottomrule
\end{tabular}
\end{table*}

\subsection{Adversarial Attacks in BCIs}

Attacking the machine learning models in BCIs could cause significant damages, ranging from user frustration to serious injuries. For example, in seizure treatment, attacks to RNS's \cite{Geller2018} seizure recognition algorithm may quickly drain its battery or make it completely ineffective, significantly reducing the patient's quality-of-life. Adversarial attacks to an EEG-based BCI speller may hijack the user's true inputs and output wrong letters, leading to user frustration or misunderstanding. In BCI-based driver drowsiness estimation \cite{drwuTFS2017}, adversarial attacks may make a drowsy driver look alert, increasing the risk of accidents.

Although most BCI research so far focused on making BCIs faster and more accurate, pioneers in BCIs have started to consider neurosecurity. For example, Ienca \emph{et al.} \cite{Ienca2018} pointed in a Nature Biotechnology Commentary in 2018 that ``\emph{greater safeguards are needed to address the personal safety, security and privacy risks arising from increasing adoption of neurotechnology in the consumer realm.}"  Judy Illes pointed out in a Nature Biotechnology Focus article \cite{Jarchum2019} in 2019 three concerns on the widespread use of brain recording and stimulation. The second is ``\emph{devices getting hacked and, by extension, behavior unwillfully and unknowingly manipulated for nefarious purposes (although this could conceivably lead to checked bad behavior too)}." A 2020 RAND Corporation report \cite{RAND2020} pointed out that ``\emph{hacking BCI capabilities could theoretically provide adversaries with direct pathways into the emotional and cognitive centers of operators' brains to sow confusion or emotional distress. In the extreme, adversary hacking into BCI devices that influence the motor cortex of human operators could theoretically send false directions or elicit unintended actions, such as friendly fire, although such influence may be technically difficult to achieve in the near term. Even an attack that broadly degraded gross motor skills could prove debilitating during combat.}" In fact, as introduced below, adversarial attacks to EEG-based BCIs have become more and more practical.

\begin{figure}[htbp]         \centering
\includegraphics[width=.8\linewidth,clip]{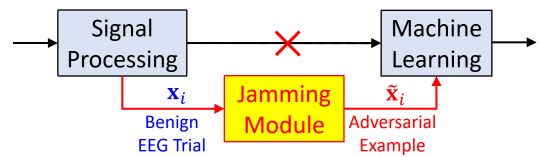}
    \caption{The BCI evasion attack approach proposed in \cite{drwuBCIAttack2019}. A jamming module is inserted between signal preprocessing and machine learning to generate adversarial examples.} \label{fig:Jamming}
\end{figure}

\begin{figure}[htbp]         \centering
\includegraphics[width=\linewidth,clip]{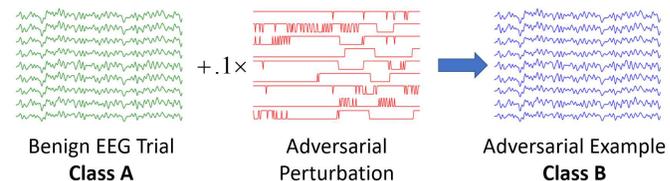}
    \caption{Evasion attack in BCIs \cite{drwuBCIAttack2019}. } \label{fig:example}
\end{figure}

In 2019, Zhang and Wu \cite{drwuBCIAttack2019} first pointed out that adversarial examples exist in EEG-based BCIs, i.e., deep learning models in BCIs are vulnerable to adversarial attacks. They successfully performed white-box, gray-box and black-box non-targeted evasion attacks to three CNN classifiers, i.e., EEGNet \cite{EEGNet}, DeepCNN and ShallowCNN \cite{Schirrmeister2017}, in three different BCI paradigms, i.e., P300 evoked potential detection, feedback error-related negativity detection, and motor imagery classification. The basic idea, shown in Fig.~\ref{fig:Jamming}, is to add a jamming module between EEG signal processing and machine learning to generate adversarial examples, optimized by unsupervised FGSM. The generated adversarial perturbations are too small to be noticed by human eyes (an example is shown in Fig.~\ref{fig:example}), but can significantly reduce the classification accuracy.

It is important to note that the jamming module is implementable, as research \cite{Sundararajan2017} has shown that BtleJuice, a framework to perform Man-in-the-Middle attacks on Bluetooth devices, can be used to intercept the data from a consumer grade EEG-based BCI system, modify them, and then send them back to the headset. The RAND report \cite{RAND2020} also pointed out that ``\emph{in a battlefield situation, these weak signals (electrical signals in BCIs) could potentially be jammed.}"

\begin{figure}[htbp]         \centering
\includegraphics[width=.98\linewidth,clip]{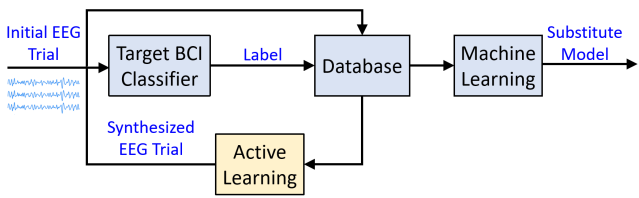}
    \caption{Query synthesis based active learning in black-box evasion attack to EEG-based BCIs \cite{drwuALBCI2019}.} \label{fig:AL}
\end{figure}

Jiang \emph{et al.} \cite{drwuALBCI2019} focused on black-box non-targeted evasion attacks to deep learning models in BCI classification problems, in which the attacker trains a substitute model to approximate the target model, and then generates adversarial examples from the substitute model to attack the target model. Learning a good substitute model is critical to the success of black-box attacks, but it requires a large number of queries to the target model. Jiang \emph{et al.} \cite{drwuALBCI2019} proposed a novel query synthesis based active learning framework to improve the query efficiency, by actively synthesizing EEG trials scattering around the decision boundary of the target model, as shown in Fig.~\ref{fig:AL}. Compared with the original black-box attack approach in \cite{drwuBCIAttack2019}, the active learning based approach can improve the attack success rate with the same number of queries, or, equivalently, reduce the number of queries to achieve a desired attack performance. This is the first work that integrates active learning and adversarial attacks for EEG-based BCIs.

The above two studies considered classification problems, as in most adversarial attack research. Adversarial attacks to regression problems were much less investigated in the literature. Meng \emph{et al.} \cite{drwuTAR2019} were the first to study white-box targeted evasion attacks for BCI regression problems. They proposed two approaches, based on optimization and gradient, respectively, to design small perturbations to change the regression output by a pre-determined amount. Experiments on two BCI regression problems (EEG-based driver fatigue estimation, and EEG-based user reaction time estimation in the psychomotor vigilance task) verified their effectiveness: both approaches can craft adversarial EEG trials indistinguishable from the original ones, but can significantly change the outputs of the BCI regression model. Moreover, adversarial examples generated from both approaches are also transferable, i.e., adversarial examples generated from one known regression model can also be used to attack an unknown regression model in black-box settings.

The above three attack strategies are theoretically important, but there are some constraints in applying them to real-world BCIs:
\begin{enumerate}
\item \emph{Trial-specificity}, i.e., the attacker needs to generate different adversarial perturbations for different EEG trials.
\item \emph{Channel-specificity}, i.e., the attacker needs to generate different adversarial perturbations for different EEG channels.
\item \emph{Non-causality}, i.e., the complete EEG trial needs to be known in advance to compute the corresponding adversarial perturbation.
\item \emph{Synchronization}, i.e., the exact starting time of the EEG trial needs to be known for the best attack performance.
\end{enumerate}
Some recent studies tried to overcome these constraints.

\begin{figure}[htbp]         \centering
\includegraphics[width=\linewidth,clip]{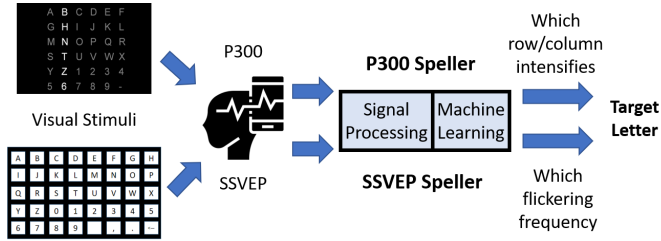}
\caption{Workflow of a P300 speller and an SSVEP speller \cite{drwuNSR2021}. For each speller, the user watches the stimulation interface, focusing on the character he/she wants to input, while EEG signals are recorded and analyzed by the speller. The P300 speller first identifies the row and the column that elicit the largest P300, and then outputs the letter at their intersection. The SSVEP speller identifies the output letter directly by matching the user’s EEG oscillation frequency with the flickering frequency of each candidate letter.} \label{fig:Speller}
\end{figure}

Zhang \emph{et al.} \cite{drwuNSR2021} performed white-box targeted evasion attacks to P300 and SSVEP based BCI spellers (Fig.~\ref{fig:Speller}), and showed that a tiny perturbation to the EEG trial can mislead the speller to output any character the attacker wants, e.g., change the output from `Y' to `N', or vice versa. The most distinguishing characteristic of their approach is that it explicitly considers the causality in designing the perturbation, i.e., it should be generated before or as soon as the target EEG trial starts, so that it can be added to the EEG trial in real-time in practice. To achieve this, an adversarial perturbation template is constructed from the training set only and then fixed. So, there is no need to know the test EEG trial and compute the perturbation specifically for it. Their approach resolves the trial-specificity and non-causality constraints, but different EEG channels still need different perturbations, and it also requires the attacker to know the starting time of an EEG trial in advance to achieve the best attack performance, i.e., there are still channel-specificity and synchronization constraints.

Zhang \emph{et al.} \cite{drwuNSR2021} considered targeted attacks to a traditional and most frequently used BCI speller pipeline, which has separate feature extraction and classification steps. Liu \emph{et al.} \cite{drwuUAP2021} considered both targeted and non-targeted white-box evasion attacks to end-to-end deep learning models in EEG-based BCIs, and proposed a total loss minimization (TLM) approach to generate universal adversarial perturbations (UAPs) for them. Experimental results demonstrated its effectiveness on three popular CNN classifiers (EEGNet, ShallowCNN, and DeepCNN) in three BCI paradigms (P300, feedback error related negativity, and motor imagery). They also verified the transferability of UAPs in non-targeted gray-box evasion attacks.

To further simplify the implementation of TLM-UAP, Liu \emph{et al.} \cite{drwuUAP2021} also considered smaller template size, i.e., mini TLM-UAP with a small number of channels and time domain samples, which can be added anywhere to an EEG trial. Mini TLM-UAPs are more practical and flexible, because they do not require the attacker to know the exact number of EEG channels and the exact length and starting time of an EEG trial. Liu \emph{et al.} \cite{drwuUAP2021} showed that, generally, all mini TLM-UAPs were effective. However, their effectiveness decreased when the number of used channels and/or the template length decrease, which is intuitive. This is the first study on UAPs of CNN classifiers in EEG-based BCIs, and also the first on optimization based UAPs for targeted evasion attacks.

In summary, the TLM-UAP approach \cite{drwuUAP2021} resolves the trial-specificity and non-causality constraints, and mini TLM-UAPs further alleviate the channel-specificity and synchronization constraints.

\begin{figure}[htbp]         \centering
\includegraphics[width=\linewidth,clip]{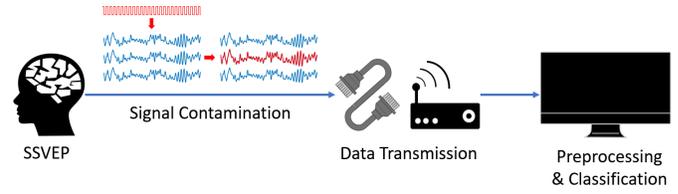}
\caption{Square wave evasion attack to SSVEP spellers \cite{drwuSCIS2022}.} \label{fig:SW}
\end{figure}

More recently, Bian \emph{et al.} \cite{drwuSCIS2022} proposed square wave evasion attacks to two popular training-free models (canonical correlation analysis, and filter-bank canonical correlation analysis) in SSVEP-based BCI spellers. As shown in Fig.~\ref{fig:SW}, the attacker only needs to know the frequency of the target character; by adding a square wave of that frequency to any input SSVEP trial, the output character can be changed to the target character with almost 100\% success rate. The proposed attack approach can resist EEG preprocessing, is robust to SSVEP trial length, and is insensitive to the phase of the square wave signal, i.e., the attacker can use any random initial phase. This represents so far the easiest implementation of evasion attacks to SSVEP-based BCI systems.

\begin{figure}[htbp]         \centering
\includegraphics[width=\linewidth,clip]{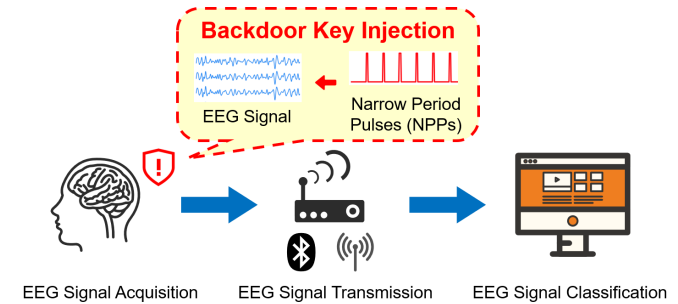}
\caption{Poisoning attack in EEG-based BCIs \cite{drwuEngineering2022}. Narrow period pulses can be added to EEG trials during signal acquisition.} \label{fig:NPP}
\end{figure}

All above studies focused on evasion attacks. Meng \emph{et al.} \cite{drwuEngineering2022} were the first to show that poisoning attacks can also be performed for EEG-based BCIs, as shown in Fig.~\ref{fig:NPP}. They proposed a practically realizable backdoor key, narrow period pulse, for EEG signals, which can be inserted into the benign EEG signal during data acquisition, and demonstrated its effectiveness in black-box targeted poisoning attacks, i.e., the attacker does not know any information about the test EEG trial, including its starting time, and wants to classify the test trial into a specific class, regardless of its true class. In other words, it resolves the trial-specificity, channel-specificity, causality and synchronization constraints simultaneously. To our knowledge, this is to-date the most practical BCI attack approach.

A summary of existing adversarial attack approaches in EEG-based BCIs is shown in Table~\ref{tab:comp2}.

\begin{table}[htbp] \centering \setlength{\tabcolsep}{3mm}
\caption{Characteristics of existing adversarial attack approaches in EEG-based BCIs. `$\checkmark$' means that constraint is satisfied,  `$\sim$' partially resolved, and `$\times$' not satisfied.}   \label{tab:comp2}
\begin{tabular}{c|cccc} \toprule
\multirow{2}{*}{Reference} & Trial- & Channel- & Non- & Synchro- \\
 & Specificity & Specificity & Causality & nization \\ \midrule
\cite{drwuBCIAttack2019} &$\times$ &$\times$&$\times$&$\times$ \\
\cite{drwuALBCI2019} &$\times$ &$\times$&$\times$&$\times$ \\
\cite{drwuTAR2019} &$\times$ &$\times$&$\times$&$\times$  \\
\cite{drwuNSR2021} & $\checkmark$ & $\times$& $\checkmark$ & $\times$ \\
\cite{drwuUAP2021}& $\checkmark$ & $\sim$& $\checkmark$ & $\sim$ \\
\cite{drwuEngineering2022} & $\checkmark$ & $\checkmark$& $\checkmark$ & $\checkmark$ \\
\cite{drwuSCIS2022} & $\checkmark$ & $\checkmark$& $\checkmark$ & $\checkmark$ \\  \bottomrule
\end{tabular}
\end{table}

\subsection{Adversarial Attacks in Health Informatics}

Adversarial attacks in health informatics can also cause serious damages, even deaths. For example, adversarial attacks to the machine learning algorithms in implantable cardioverter defibrillators could lead to unnecessary painful shocks, damaging the cardiac tissue, and even worse therapy interruptions and sudden cardiac death \cite{Paoletti2019}.

Han \emph{et al.} \cite{Han2020} proposed both targeted and non-targeted white-box evasion attack approaches to construct smoothed adversarial examples for ECG trials that are invisible to one board-certified medicine specialist and one cardiac electrophysiology specialist, but can successfully fool a CNN classifier for arrhythmia detection. They achieved 74\% attack success rate (74\% of the test ECGs originally classified correctly were assigned a different diagnosis, after adversarial attacks) on atrial fibrillation classification from single-lead ECG collected from the AliveCor personal ECG monitor. This study suggests that it is important to check if ECGs have been altered before using them in medical machine learning models.

Aminifar \cite{Aminifar2020} studied white-box targeted evasion attacks in EEG-based epileptic seizure detection, through UAPs. He computed the UAPs via solving an optimization problem, and showed that they can fool a support vector machine classifier to misclassify most seizure samples into non-seizure ones, with imperceptible amplitude.

Newaz \emph{et al.} \cite{Newaz2020} investigated adversarial attacks to machine learning-based smart healthcare systems, consisting of 10 vital signs, e.g., EEG, ECG, SpO$_2$, respiration, blood pressure, blood glucose, blood hemoglobin, etc. They performed both targeted and non-targeted attacks, and both poisoning and evasion attacks. For evasion attacks, they also considered both white-box and black-box attacks. They showed that adversarial attacks can significantly degrade the performance of four different classifiers in smart health system in detecting diseases and normal activities, which may lead to to erroneous treatment.

Deep learning has been extensively used in health informatics; however, generally it needs a large amount of training data for satisfactory performance. Transfer learning \cite{drwuTLBCI2022} can be used to alleviate this requirement, by making use of data or machine learning models from an auxiliary domain or task. Wang \emph{et al.} \cite{Wang2020} studied targeted backdoor attacks against transfer learning with pre-trained deep learning models on both image and time series (e.g., ECG). Three optimization strategies, i.e., ranking-based neuron selection, autoencoder-powered trigger generation and defense-aware retraining, were used to generate backdoors and retrain deep neural networks, to defeat pruning based, fine-tuning/retraining based and input pre-processing based defenses. They demonstrated their effectiveness in brain MRI image classification and ECG heartbeat type classification.

\subsection{Adversarial Attacks in Biometrics}

Physiological signals, e.g., EEG, ECG and PPG, have recently been used in biometrics \cite{Singh2012}. However, they are subject to presentation attacks in such applications. In a physiological signal based presentation attack, the attacker tries to spoof the biometric sensors with a fake piece of physiological signal \cite{Eberz2017}, which would be authenticated as from a specific victim user.

Maiorana \emph{et al.} \cite{Maiorana2013} investigated the vulnerability of an EEG-based biometric system to hill-climbing attacks. They assumed that the attacker can access the matching scores of the biometric system, which can then be used to guide the generation of synthetic EEG templates until a successful authentication is achieved. This is essentially a black-box targeted evasion attack in the adversarial attack terminology: the synthetic EEG signal is the adversarial example, and the victim's identify is the target class. It's a black-box attack, because the attacker can only observe the output of the biometric system, but does not know anything else about it.

\begin{figure*}[htpb]\centering
\subfigure[]{ \label{fig:offlineECG}   \includegraphics[width=.8\linewidth,clip]{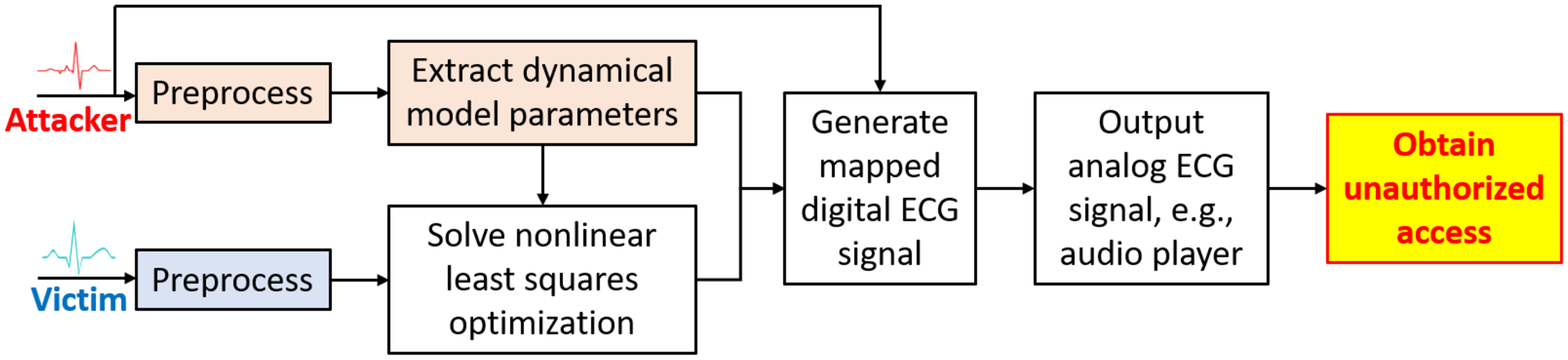}}
\subfigure[]{ \label{fig:onlineECG}    \includegraphics[width=.85\linewidth,clip]{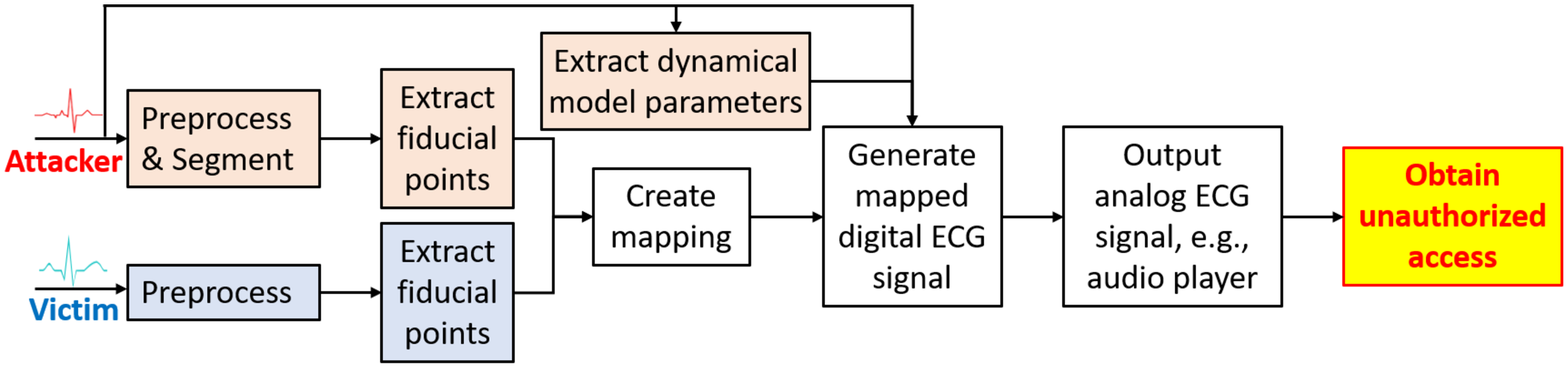}}
\caption{(a) Offline ECG biometrics presentation attack \cite{Eberz2017}; (b) Online ECG biometrics presentation attack \cite{Karimian2017}. } \label{fig:ECG}
\end{figure*}

Eberz \emph{et al.} \cite{Eberz2017} proposed an offline ECG biometrics presentation attack approach, illustrated in Fig.~\ref{fig:offlineECG}. The basic idea was to find a mapping function to transform ECG trials recorded from the attacker so that they resemble the morphology of ECG trials from a specific victim. The transformed ECG trials can then be used to fool an ECG biometric system to obtain unauthorized access. They showed that the attacker ECG trials can be obtained from a device different from the one that the victim ECG trials are recorded from (i.e., cross-device attack), and there could be different approaches to present the transformed ECG trials to the biometric device under attack, the simplest being the playback of ECG trials encoded as .wav files using an off-the-shelf audio player.

Unlike \cite{Maiorana2013}, the above approach is a gray-box targeted evasion attack in the adversarial attack terminology: the attacker's ECG signal can be viewed as the benign example, the transformed ECG signal is the adversarial example, and the victim's identity is the target class. The mapping function plays the role of the jamming module in Fig.~\ref{fig:Jamming}. It's a gray-box attack, because the attacker needs to know the feature distributions of the victim ECGs in designing the mapping function.

Karimian \emph{et al.} \cite{Karimian2020} proposed an online ECG biometrics presentation attack approach, shown in Fig.~\ref{fig:onlineECG}. Its procedure is very similar to the offline attack one in Fig.~\ref{fig:offlineECG}, except that the online approach is simpler, because it only requires as few as one victim ECG segment to compute the mapping function, and the mapping function is linear. Karimian \cite{Karimian2019} also proposed a similar presentation attack approach to attack PPG-based biometrics. Again, these approaches can be viewed as gray-box targeted evasion attacks.

More recently, Karim \emph{et al.} \cite{Karim2021} utilized an adversarial transformation network on a distilled model to attack two classification models (1-nearest neighbor dynamic time warping, and a fully convolutional network) on 42 time series datasets. There were three ECG datasets on the classification of
humans with heart conditions or Myocardial infarctions, and two EOG datasets on the classification of Japanese Katakana strokes. They performed both white-box and black-box non-targeted evasion attacks. However, they did not consider the causality of the time-series, i.e., the entire test trial was used in generating the adversarial perturbation. So, their approach may only be used offline.

\subsection{Discussion} \label{sect:discussion}

Although we have not found adversarial attack studies on affective computing and adaptive automation in physiological computing, it does not mean that adversarial attacks cannot be performed in such applications. Machine learning models in affective computing and adaptive automation are not fundamentally different from those in BCIs; so, adversarial attacks in BCIs can easily be adapted to affective computing and adaptive automation. Particularly, Meng \emph{et al.} \cite{drwuTAR2019} have shown that it is possible to attack the regression models in EEG-based driver fatigue estimation and EEG-based user reaction time estimation, whereas driver fatigue and user reaction time could be triggers in adaptive automation.

It is interesting to note that almost all aforementioned adversarial attacks focused on EEG and ECG, the 2nd and 3rd most popular physiological signals in Table~\ref{tab:pubs}. Blood pressure, the most popular physiological signal, was not attacked alone; it was considered only once in \cite{Newaz2020}, together with EEG, ECG, etc. The reason may be that blood pressure consists of only two numbers (systolic and diastolic pressures) measured infrequently, so it is not easy and interesting to attack. The same reasoning may also apply to respiration and heart rate. Some other physiological signals, e.g., ECoG and EMG, are frequently used in human-machine interactions, and also may be complex and important enough to attract adversarial attacks.

Finally, examining the current few studies on adversarial attacks of time series \cite{Fawaz2019,Karim2021,Harford2021}, we found that all of them did not take the causality of the time series into consideration, i.e., their approaches utilized the entire test trial in computing the adversarial perturbation, so they can only be used offline. Additionally, they used generic classifiers for all time series, whereas in physiological computing, particularly BCIs \cite{drwuTLBCI2022}, each paradigm has its own best feature extraction and classification/regression approach, according to the neurological basis of the corresponding paradigm. So, these generic time series attack approaches may not be used directly in physiological computing.

\section{Defense Against Adversarial Attacks} \label{sect:defense}

There are different adversarial defense strategies \cite{Qiu2019,Bernal2021}:
\begin{enumerate}
\item \emph{Data modification}, which modifies the training set in the training stage or the input data in the test stage, through adversarial training \cite{Szegedy2014}, gradient hiding \cite{Tramer2017}, transferability blocking \cite{Hosseini2017}, data compression \cite{Das2017}, data randomization \cite{Xie2017}, etc.

\item \emph{Model modification}, which modifies the target model directly to increase its robustness. This can be achieved through regularization \cite{Biggio2011}, defensive distillation \cite{Papernot2016}, feature squeezing \cite{Xu2017},  using a deep contractive network \cite{Gu2014} or a mask layer \cite{Gao2017}, etc.

\item \emph{Auxiliary tools}, which may be additional auxiliary machine learning models to robustify the primary model, e.g., adversarial detection models \cite{Qayyum2021}, or defense generative adversarial nets (defense-GAN) \cite{Samangouei2018}, high-level representation guided denoiser \cite{Liao2018}, etc.
\end{enumerate}

As researchers just started to investigate adversarial attacks in physiological computing, there were even fewer studies on defense strategies against them. A summary of them is shown in Table~\ref{tab:comp3}.

\begin{table}[htbp] \centering \setlength{\tabcolsep}{1mm}
\caption{Summary of existing adversarial defense studies in physiological computing.}   \label{tab:comp3}
\begin{tabular}{c|cccc} \toprule
\multirow{2}{*}{Reference} & \multirow{2}{*}{Application} & Data  & Model  & Adversarial  \\
 &  &  Modification & Modification &  Detection \\ \midrule
\cite{Hussein2020} &BCI & $\checkmark$ && \\
\cite{Sadeghi2019} &BCI & &$\checkmark$& \\
\cite{Cai2015} &Health Informatics &&&$\checkmark$ \\
\cite{Cai2016}&Health Informatics &&&$\checkmark$ \\
\cite{Karim2021} & Health Informatics &$\checkmark$&&  \\
\cite{Karimian2020} & Biometrics &&&$\checkmark$  \\
\bottomrule
\end{tabular}
\end{table}

\subsection{Adversarial Training}

Adversarial training, which trains a robust machine learning model on normal plus adversarial examples, may be the most popular data modification based adversarial defense approach.

Hussein \emph{et al.} \cite{Hussein2020} proposed an approach to augment deep learning models with adversarial training for robust prediction of epilepsy seizures. Though their goal was to overcome some challenges in EEG-based seizure classification, e.g., individual differences and shortage of pre-ictal labeled data, their approach can also be used to defend against adversarial attacks.

They first constructed a deep learning classifier from available limited amount of labeled EEG data, and then performed white-box attacks to the classifier to obtain adversarial examples, which were next combined with the original labeled data to retrain the deep learning classifier. Experiments on two public seizure datasets demonstrated that adversarial training increased both the classification accuracy and the classifier robustness.

More recently, Karim \emph{et al.} \cite{Karim2021} performed adversarial training for fully convolutional network classifiers, and tested the performance on 42 time series datasets, including three ECG datasets on heart conditions or Myocardial infarction classification, and two EOG datasets on Japanese Katakana stroke classification. They showed that even very simple adversarial training can improve the robustness of fully convolutional network classifiers to black-box and white-box non-targeted evasion attacks.

Although adversarial training may be the most effective approach for enhancing the robustness of a model, it could lead to undesirable accuracy degradation on the benign examples \cite{Rade2022}. Additionally, it increases the computational cost 3-30 times \cite{Shafahi2019}.

\subsection{Model Modification}

Regularization based model modification to defend against adversarial attacks usually considers the model security (robustness) in the optimization objective function.

Sadeghi \emph{et al.} \cite{Sadeghi2019} proposed an analytical framework for tuning the classifier parameters, to ensure simultaneously its accuracy and security. The optimal classifier parameters were determined by solving an optimization problem, which takes into account both the test accuracy and the robustness against adversarial attacks. For $k$-nearest neighbor (kNN) classifiers, the two parameters to be optimized are the number of neighbors and the distance metric type. Experiments on EEG-based eye state (open or close) recognition verified that it is possible to achieve both high classification accuracy and high robustness against black-box targeted evasion attacks.

Model modification approaches are usually heuristic and empirical, without theoretical guarantees. They may be vulnerable to model-agnostic block-box attacks \cite{Carlini2017}.

\subsection{Adversarial Detection}

Adversarial detection uses a separate module to detect if there is adversarial attack, and takes actions accordingly. The simplest is to discard adversarial examples directly.

Cai and Venkatasubramanian \cite{Cai2016} proposed an approach to detect signal injection-based morphological alterations (evasion attack) of ECGs. Because multiple physiological signals based on the same underlying physiological process (e.g., cardiac process) are inherently related to each other, any adversarial alteration of one of the signals will lead to inconsistency in the other signal(s) in the group. Since both ECG and arterial blood pressure measurements are representations of the cardiac process, the latter can be used to detect morphological alterations in ECGs. They demonstrated over 90\% accuracy in detecting even subtle ECG morphological alterations for both healthy subjects and patients. A similar idea \cite{Cai2015} was also used to detect temporal alternations of ECGs, by making use of their correlations with arterial blood pressure and respiration measurements.

Karimian \emph{et al.} \cite{Karimian2020} proposed two strategies to protect ECG biometric authentication systems from spoofing, by evaluating if ECG signal characteristics match the corresponding heart rate variability or PPG features (pulse transit time and pulse arrival time). The idea is actually similar to Cai and Venkatasubramanian's \cite{Cai2016}. If there is a mismatch, then the system considers the input to be fake, and rejects it.

Adversarial detection heavily relies on the difference between adversarial examples and benign examples. However, adversarial examples can fool not only the classifier but also the detector, so adversarial detection may be ineffective against adaptive attacks \cite{Carlini2017a}.

\subsection{Discussion}

Although there have been many adversarial attack defense approaches \cite{Miller2020}, no one can withstand all existing attacks, not to mention new attacks that will for sure be discovered in the future. For example, Miller \emph{et al.}'s experiments \cite{Miller2020} showed that an adversarially trained robust deep neural network can accommodate small perturbations, at the cost of significant classification accuracy loss (about 10\%) for benign inputs. Furthermore, as the attack strength increased, this robust classifier gradually became ineffective.

As proposed in \cite{Miller2020}, a promising new adversarial defense direction may be to combine robust classification with detection: robust classification forces the adversarial perturbations to be large for successful attacks, but this also makes the attacks more detectable. So, using robust classification and adversarial detection together may outperform each one alone.

Another idea is to use multi-modal inputs in physiological computing. Multi-modal signals are frequently used to increase the accuracy of physiological computing, e.g., using both EEG and EOG increased the emotion classification accuracy significantly \cite{Zheng2019a}. They can also be used to increase the robustness to adversarial attacks, as different physiological signals generally require different perturbations, which raises the difficulty of attacking. However, this may also increase the complexity and cost of the resulting physiological computing system. Thus, there should be a careful trade-off between accuracy/robustness and complexity/cost.

\section{Conclusions and Future Research} \label{sect:conclusion}

Physiological computing includes, or significantly overlaps with, BCIs, affective computing, adaptive automation, health informatics, and physiological signal based biometrics. It increases the communication bandwidth from the user to the computer, but is also subject to adversarial attacks. This paper has given a comprehensive review on adversarial attacks and their defense strategies in physiological computing, hopefully will bring more attention to the security of physiological computing systems.

Promising future research directions in this area include:
\begin{enumerate}
\item Transfer learning has been extensively used in physiological computing \cite{drwuTLBCI2022}, to alleviate the training data shortage problem by leveraging data from other subjects \cite{drwuEA2020} or tasks \cite{drwuTNSRE2016}, or to warm-start the training of a (deep) learning algorithm by borrowing parameters or knowledge from an existing algorithm \cite{Wang2020}, as shown in Fig.~\ref{fig:TL}. However, transfer learning is particularly susceptive to poisoning attacks \cite{drwuEngineering2022,Wang2020}. It's very important to develop strategies to check the integrity of data and models before using them in transfer learning.

\begin{figure}[htbp]         \centering
\includegraphics[width=.95\linewidth,clip]{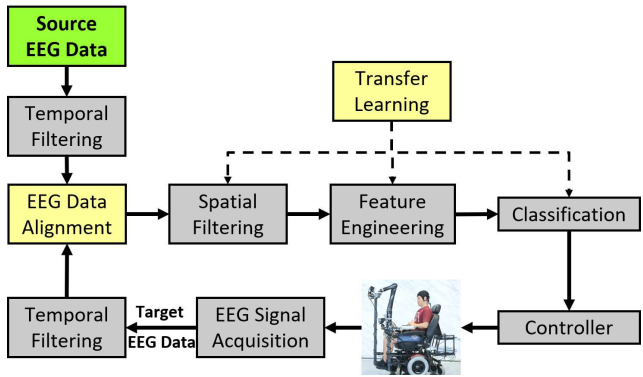}
    \caption{A transfer learning pipeline in motor imagery based BCIs \cite{drwuMITLBCI2022}.} \label{fig:TL}
\end{figure}

\item Adversarial attacks to other components in the machine learning pipeline (an example on BCI is shown in Fig.~\ref{fig:attack}), which includes signal processing, feature engineering, and classification/regression, and the corresponding defense strategies. So far all adversarial attack approaches in physiological computing considered the classification or regression model only, but not other components, e.g., signal processing and feature engineering. It has been shown that feature selection is also subjective to data poisoning attacks \cite{Xiao2015a}, and adversarial feature selection can be used to defend against evasion attacks \cite{Zhang2016a}.

\begin{figure}[htbp]         \centering
\includegraphics[width=.95\linewidth,clip]{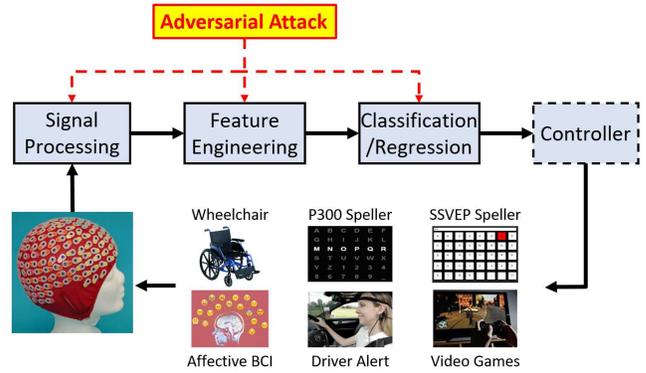}
    \caption{Adversarial attacks to the BCI machine learning pipeline.} \label{fig:attack}
\end{figure}

\item Additional types of attacks in physiological computing \cite{Denning2009,Rushanan2014,Camara2015,Pycroft2016,Bernal2021}, and the corresponding defense strategies, as shown in Fig.~\ref{fig:addAttack}. For example, Paoletti \emph{et al.} \cite{Paoletti2019} performed parameter tampering attacks on Boston Scientific implantable cardioverter defibrillators, which use a discrimination tree to detect tachycardia episodes and then initiate the appropriate therapy. They slightly modified the parameters of the discrimination tree to achieve both attack effectiveness and stealthiness. These attacks are also very dangerous in physiological computing, and hence deserve adequate attention.

\begin{figure}[htbp]         \centering
\includegraphics[width=.8\linewidth,clip]{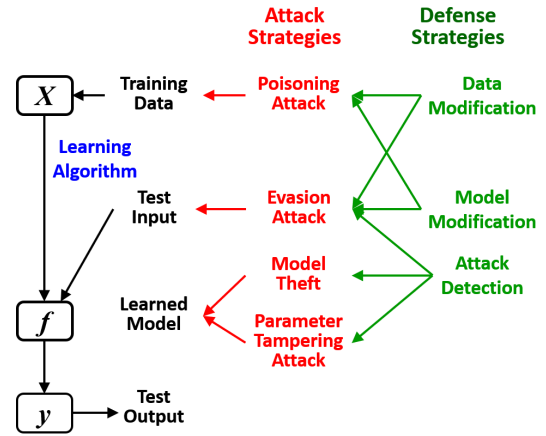}
    \caption{Additional types of attacks in physiological computing.} \label{fig:addAttack}
\end{figure}

\item Adversarial attacks to affective computing and adaptive automation applications, which have not been studied yet, but are also possible and dangerous. Many existing attack approaches in BCIs, health informatics and biometrics can be extended to them, either directly or with slight modifications. However, there could also be unique attack approaches specific to these areas. For example, emotions are frequently represented as continuous numbers in the 3D space of valence, arousal and dominance in affective computing \cite{Mehrabian1980}, and hence adversarial attacks to regression models in affective computing should be paid enough attention to.

\item Real-world demonstration of adversarial attacks and defenses. As mentioned in Section~\ref{sect:discussion}, current research on adversarial attacks of time series did not consider their causality, so the attacks may not be used in the most meaningful online applications. Adversarial attacks to BCIs have advanced rapidly in the past few years, and the attacks have become very easy to perform in theory. However, real-world experiments are still needed to demonstrate their practicableness, and more importantly, the necessity, feasibility and benefits of adversarial defenses.

\item Privacy of physiological computing systems. Adversarial attacked discussed in this review focused on manipulating the system outputs; however, privacy is another very important concern in physiological computing. For example, personal account, personal preferences, physical state and commercial models are privacy information that could be stolen from BCIs \cite{drwuTCSS2022}. Defending against these privacy attacks is also crucial for wide-spread applications of physiological computing systems.

\end{enumerate}

Finally, we need to emphasize that the goal of adversarial attack research in physiological computing should be discovering its vulnerabilities, and then finding solutions to make it more secure, instead of merely causing damages to it.\\

\end{document}